# Swarm-in-Blocks: Simplifying Drone Swarm Programming with Block-Based Language

Agnes Bressan de Almeida
João Aires Correa Fernandes Marsicano
*University of São Paulo*
São Carlos, Brazil
joao.aires.marsicano@usp.br

*Abstract*—Swarm in Blocks, originally developed for CopterHack 2022, is a high-level interface that simplifies drone swarm programming using a block-based language. Building on the Clover platform, this tool enables users to create functionalities like loops and conditional structures by assembling code blocks. In 2023, we introduced Swarm in Blocks 2.0, further refining the platform to address the complexities of swarm management in a user-friendly way.

As drone swarm applications grow in areas like delivery, agriculture, and surveillance, the challenge of managing them, especially for beginners, has also increased. The Atena team developed this interface to make swarm handling accessible without requiring extensive knowledge of ROS or programming. The block-based approach not only simplifies swarm control but also expands educational opportunities in programming.

## I. INTRODUCTION

Swarm robotics is a field of multi-robot systems where a large number of relatively simple robots cooperate to achieve complex tasks. Inspired by natural systems such as ant colonies and bird flocks, swarm robotics emphasizes the importance of decentralized control and local interactions among robots [5]. The Swarm-in-Blocks project introduces a high-level interface based on block programming, designed to make drone swarm programming accessible and straightforward, even for users with limited technical knowledge [1].

## II. PROJECT BACKGROUND

Swarm-in-Blocks originated as a project for CopterHack 2022, designed as a high-level interface using block language to simplify the programming of drone swarms. Each script in this language represents functionalities such as conditional structures, loops, or functions that handle swarm instructions. The platform is built on existing systems from COEX, specifically Clover. In 2023, Swarm-in-Blocks evolved into Swarm-in-Blocks 2.0, aiming to tackle more complex challenges associated with swarm robotics in an accessible and polished manner [2].

## III. SYSTEM ARCHITECTURE

The Swarm-in-Blocks framework provides a secure and scalable platform for swarm robotics, consisting of the following components:

- **Block-Based Programming Interface:** A user-friendly, high-level programming interface that allows users to create scripts by fitting together code blocks like puzzle pieces.
- **Robotic Nodes:** Drones equipped with sensors and communication modules to interact with the programming interface and each other.
- **Control System:** Manages the execution of scripts and coordination of drone movements based on the programmed blocks.
- **User Interface:** Tools for monitoring and managing the swarm.

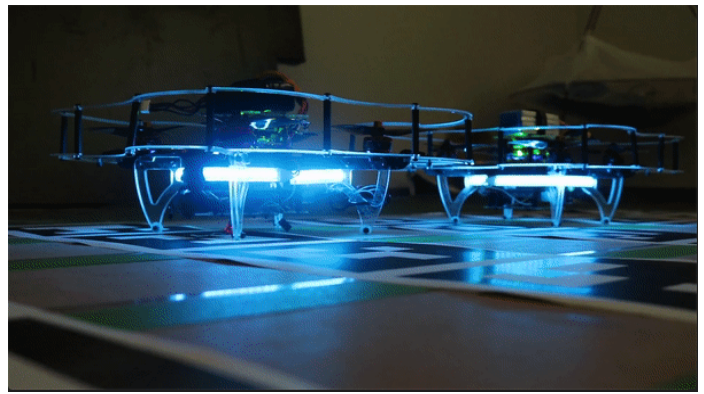

Fig. 1. Simple demo of the platform running.

### A. Block-Based Programming Interface

The block-based programming interface is designed to be intuitive, enabling users to create complex swarm behaviors without needing to write traditional code. The interface includes blocks for various functions such as movement commands, conditional logic, and sensor data processing. This approach lowers the barrier to entry for programming drone swarms, making it accessible to a wider audience [3].

### B. Robotic Nodes

Each drone in the Swarm-in-Blocks system acts as a robotic node, equipped with sensors for navigation and communication modules to interact with other drones and the control system. The drones are designed to be robust and capable

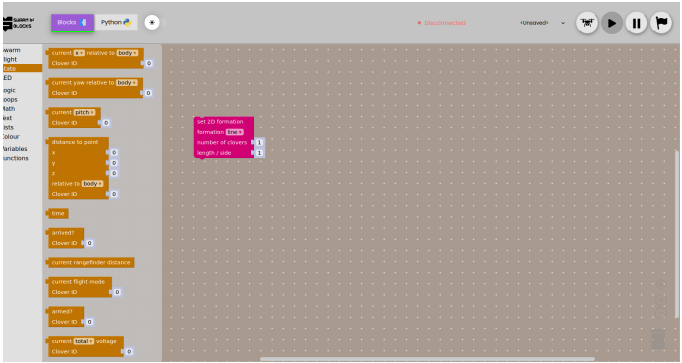

Fig. 2. Swarm Clover Blocks example.

of operating in diverse environments, from indoor spaces to outdoor fields [4].

*C. Control System*

The control system is responsible for executing the scripts created using the block-based programming interface. It interprets the block sequences and translates them into commands that are sent to the drones. The system also handles coordination between drones, ensuring that they work together to achieve the desired objectives [5].

*D. User Interface*

The user interface provides tools for monitoring and managing the swarm. Users can view the status of individual drones, track their movements in real-time, and adjust parameters as needed. The interface is designed to be intuitive and user-friendly, making it easy for users to manage complex swarm operations [3].

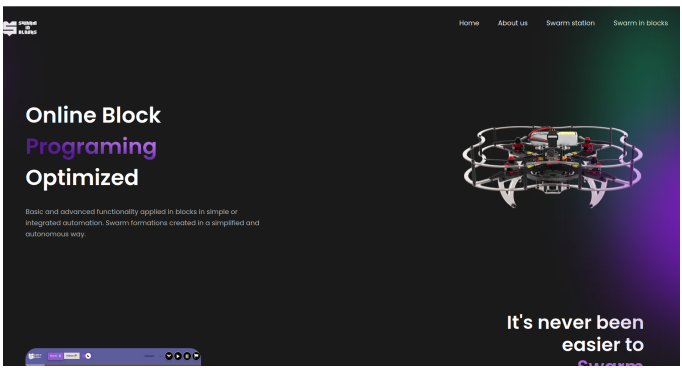

Fig. 3. Swarm in Blocks homepage

## IV. APPLICATIONS

Swarm-in-Blocks has several potential applications, including:

*A. Industrial Automation*

Swarm robotics can optimize industrial tasks like inventory management and quality control. The Swarm-in-Blocks framework enables autonomous drones to navigate warehouses, perform stocktaking, and detect damaged goods, reducing manual labor and enhancing efficiency [4].

*B. Agriculture*

In agriculture, drone swarms can handle precision tasks such as crop monitoring, pest control, and irrigation management. The block-based interface simplifies the setup of these operations, improving resource use and crop yields through targeted interventions [5].

*C. Entertainment*

Swarm-in-Blocks facilitates the creation of aerial light shows by simplifying the programming of drone choreographies. The intuitive interface allows artists to design and execute captivating performances with ease [3].

## V. BLOCK PROGRAMMING FOR GENERAL ROBOTICS

The "Swarm-in-Blocks" project, developed for the CopterHack competition and enhanced in 2023 with Swarm-in-Blocks 2.0, offers a block-based programming interface for controlling drone swarms. This project not only advances swarm control technology but also holds significant educational potential, particularly as a teaching tool.

Block-based programming, widely used in educational robotics, simplifies the learning of programming. Platforms like Blockly and Scratch allow users to stack blocks representing code segments, making it easier for students of all ages to learn coding and computational thinking skills.

*A. Educational Robotics*

Educational robotics uses robots to facilitate learning in STEM education. Tools like Scratch and Blockly make programming accessible to young learners by allowing them to drag and drop blocks representing commands and logic structures [8].

*B. Simplifying the Learning of Programming Concepts*

Block programming makes learning programming concepts intuitive and accessible. It is particularly effective for introducing students to programming without the complexity of traditional coding languages.

*C. Promoting Critical Thinking and Problem-Solving*

Programming drones in formation fosters critical thinking and problem-solving skills. Students plan, test, and adjust instructions, considering variables like speed and altitude, enhancing their logical and systematic thinking.

*D. Multidisciplinary Integration*

Swarm-in-Blocks integrates concepts from mathematics, physics, computer science, and engineering. This multidisciplinary approach helps students apply theoretical knowledge to practical situations, essential in STEM education.

*4. Preparation for the Job Market*

By introducing drone programming and blockchain technology, Swarm-in-Blocks prepares students for future job markets. Practical experience in these areas offers a competitive edge in robotics and blockchain careers.

*5. Applications in Educational Settings*

Robots like LEGO Mindstorms and VEX Robotics are used in educational settings to teach programming and robotics. These robots work with block-based coding environments, simplifying the learning process [10], [9].

*6. Impact on Learning*

Research shows that block-based programming enhances students' understanding of programming, improves problem-solving skills, and increases engagement in learning [8], [9].

## VI. SWARM IN BLOCKS SYSTEMS IN DETAILS

Due to the complexity of the Swarm in Blocks project, it was divided into systems, which represent a macroscopic view of sets of similar functionality. In this section, we will detail the developed systems, dealing not only with the features developed, but also with the motivation and implications for this.

*A. Formations*

The "Formations" of the Swarm-in-Blocks project details the use of 2D and 3D formations for drone swarms. The Python script formation.py calculates drone positions for formations such as lines, circles, squares, triangles, cubes, pyramids, and spheres. Using the Numpy library, it generates arrays of coordinates for each drone. The swarm.py file handles the application of these formations, allowing users to select and visualize formations through a menu interface. This facilitates the creation and management of complex drone formations with ease.

*B. Transformations*

The "Transformations" section of the Swarm-in-Blocks project covers operations that allow editing the drones' arrangement and the formation itself. Using matrix operations, procedures for translation, scaling, and rotation of the entire formation are developed. These transformations adjust the current coordinates of the drones, with special care for rotation, which requires centralizing the formation at the origin of the map. The collision prevention algorithm ensures that potential collisions are avoided during transformations.

*C. Collision Avoidance*

The "Collision Avoidance" section of the Swarm-in-Blocks project details an algorithm designed to prevent drone collisions within a swarm. The algorithm addresses three scenarios: one stationary and one moving drone, both drones moving on parallel paths, and both drones moving on non-parallel paths. It utilizes matrix operations and ROS (Robot Operating System) functionalities to manage drone positions and ensure safe distances are maintained, preventing collisions during swarm operations.

*D. LED Effects*

In the Swarm-in-Blocks project, the LED effects system allows for visual enhancements of drone formations. Ready-to-use effects include options for all drones, random drones, even and odd drones, and 2D formations. These effects can be controlled via user input, offering various modes like fill, fade, flash, blink, blink fast, wipe, rainbow, and rainbow fill. Each effect alters the LED behavior to enhance visibility and aesthetics, with some effects tailored to specific formations.

*E. First Person View (FPV)*

Part of the Swarm-in-Blocks 2.0, the First Person View (FPV) system enhances user experience by allowing real-time control of drones via a camera feed. Developed using roslibjs for web integration with ROS, it enables users to view and control drones through a web interface using JavaScript, HTML, and CSS. Users can select drones, view live footage, and control drone movements with keyboard inputs, improving flight safety and control.

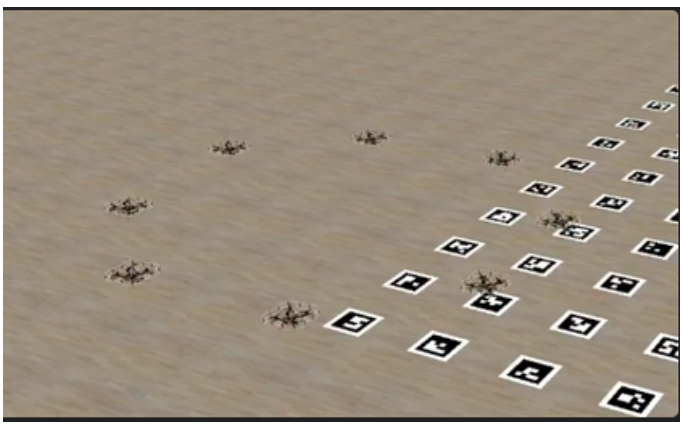

Fig. 4. Circle formation.

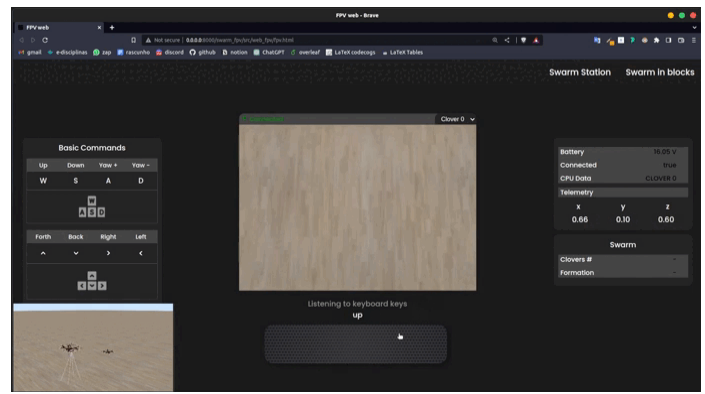

Fig. 5. Swarm in Blocks 2023 FPV.

*C. Smart Preview*

Swarm Preview in the Swarm-in-Blocks project enables users to visualize swarm behavior in simulations or real life with minimal computational power. It's used primarily in Planning Mode but also in Simulation and Navigation Modes. Users can choose 2D or 3D previews via the Python or Blocks API. Functions include resuming, saving, and canceling simulations, and navigating through operations with control arrows to identify issues.

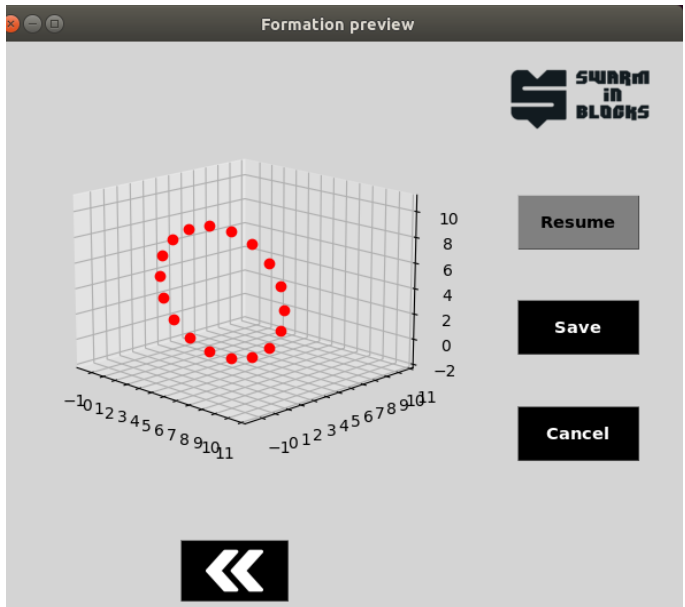

Fig. 6. 3D Plot Preview menu.

## VII. CODE REVIEW

In this section, we delve deeper into the codebase that underpins the core components of our Swarm-in-Blocks platform. The primary focus will be on examining the code responsible for the key pillars of our application: the **Swarm Clover Blocks**, the **homepage**, the **First Person View (FPV)** and the **Swarm Station**. These pillars form the foundation of our platform, each contributing unique functionality and ensuring seamless interaction between users and drone swarms.

The subsections that follow will present detailed explanations and code snippets to illustrate how these pillars are constructed, focusing on the logic and structure that enable the Swarm-in-Blocks platform to function effectively.

*A. Swarm Clover Blocks*

The `swarm_clover_blocks` is a crucial component of the Swarm-in-Blocks platform, serving as the backend that enables block-based programming for drone swarms. This backend application extends the functionalities of the original `clover_blocks` package, integrating support for controlling multiple drones simultaneously, which is essential for swarm operations.

*1) Code Structure:* The `swarm_clover_blocks` package is organized into several directories and files, each playing a specific role in the overall functionality of the platform:

*a) msg/:* This directory contains the message definitions used within the system. These messages are essentially the data structures that encapsulate user requests and responses, along with any necessary identifiers. The use of ROS messages allows for standardized communication between different nodes in the ROS ecosystem.

*b) programs/:* In this directory, pre-configured examples of block-based programs are stored. These examples are complete with the necessary blocks to perform specific tasks, providing users with ready-made templates that can be easily customized through the frontend interface.

*c) src/:* This is the core of the backend, where the primary logic of the `swarm_clover_blocks` resides. The key file here is `clover_blocks.py`, which orchestrates all backend operations, including the execution of user-defined programs and management of communication between the frontend and the drones.

*d) srv/:* The `srv/` directory defines the service files, which specify the structure of requests and responses for operations like loading, running, and storing programs. These services are essential for the dynamic interaction between the frontend interface and the backend logic.

*e) www/:* This directory forms the heart of the frontend. It includes the core Blockly files as well as custom adaptations specific to the Swarm-in-Blocks platform. Key files include `blocks.js`, which defines the available blocks and their functionalities, `index.html` for the web interface, and `main.css` for styling the interface. Additionally, `ros.js` and `roslib.js` are used for communication with the ROS backend.

*2) Node Functionality:* The `swarm_clover_blocks` node is the operational backbone of the block-based programming interface. It implements all the necessary services and topics to run Blockly-generated Python scripts. The node's functionality includes:

- **Service** `run` – Executes the Blockly-generated program, interpreting the block sequence as a Python script and commanding the drones accordingly.
- **Service** `stop` – Provides the ability to terminate an ongoing program, ensuring that drone operations can be halted safely and immediately.
- **Service** `store` – Allows users to save their block-based programs for future use, facilitating easy access and reusability of code.
- **Service** `load` – Retrieves stored programs from the `programs/` directory, making them available for execution through the frontend.

*3) Parameters and Topics:* The `swarm_clover_blocks` node also manages several parameters and topics that govern its operation:

*a) Parameters:* The parameters control various aspects of the drone's behavior, such as navigation tolerances, yaw angle precision, and confirmation prompts before running a program. These parameters can be adjusted directly in the node configuration or via URL GET-parameters when accessing the web interface.

*b) Published Topics:* The node publishes several topics, including:

- `running:` Indicates whether a program is currently being executed.
- `block:` Provides updates on the current block being executed, which is particularly useful for monitoring program progress.
- `error:` Reports any errors or exceptions that occur during program execution, ensuring that users are informed of issues in real time.
- `prompt:` Manages user input requests during program execution, allowing for interactive control of the drone swarm.

This comprehensive structure ensures that the `swarm_clover_blocks` package can effectively manage the complex task of programming and controlling drone swarms, providing both a user-friendly frontend and a powerful backend.

For more detailed information about the Swarm Clover Blocks, please refer to the official repository available at Swarm-in-Blocks Repository.

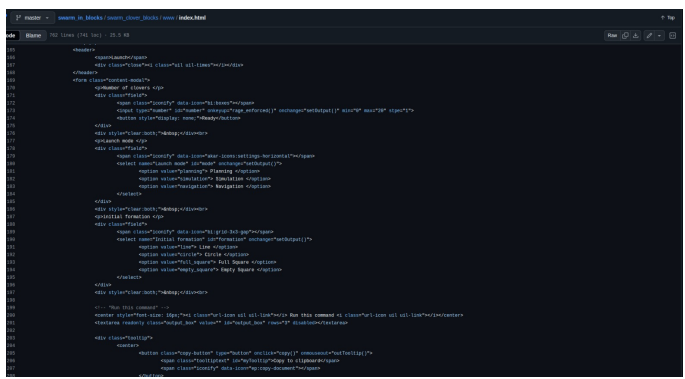

Fig. 7. Swarm Clover Blocks Code.

*B. homepage*

The `Clover_UI` serves as the main user interface for the Swarm-in-Blocks platform, providing users with a comprehensive homepage to launch and manage both individual drones and drone swarms. This interface is accessible through a web browser by navigating to the localhost address configured via Apache, making it easy to control and monitor drones directly from any device on the same network.

*1) Code Structure:* The `Clover_UI` package is organized into a ReactJS project, with a clear structure that facilitates development, deployment, and maintenance. The following are the key components of the project:

*a) `cloverUI/`:* This is the root directory of the ReactJS project. It contains all the essential files and folders needed to run the Clover UI, including the main HTML file and source code.

*b) `cloverUI/index.html`:* This file is a crucial part of the ReactJS configuration. It acts as an entry point for the application, calling the main project components but not containing the actual application code.

*c) `cloverUI/src/`:* The `src/` directory is where the core logic and algorithms of the Clover UI reside. This is the primary source folder, containing all the React components and configuration files that drive the application's functionality.

*d) `cloverUI/src/App.jsx`:* The `App.jsx` file is the heart of the Clover UI project. It orchestrates the entire application by calling all the component pages. Each specific component's code is located in the `Components` folder, and `App.jsx` brings these components together to form the full user interface.

*e) `cloverUI/src/Componentes/`:* This folder contains the essential components that make up the application. Each component represents a specific part of the user interface, such as the header, carousel, or action buttons.

*f) `cloverUI/src/style.js`:* The `style.js` file contains Vite configuration settings that apply the defined styles dynamically. Vite is used to manage the project's build process, ensuring that styles are correctly applied during both development and production builds.

*2) Frontend Technologies:* The Clover UI leverages the ReactJS framework to create a responsive and interactive user interface. ReactJS was chosen due to its popularity, fast learning curve, and active community, making it an ideal choice for developing a robust and scalable frontend.

Additionally, the Clover UI takes advantage of complex visual effects, such as the Slider Carousel, to enhance the user experience. ReactJS's flexibility and ease of use enabled the development of a platform that is both functional and user-friendly, even though the page does not have its own backend.

To further streamline the development process, Vite, a ReactJS project builder, was used. Vite simplifies the creation and configuration of ReactJS projects, providing a pre-configured environment that is ready to use. This allowed the development team to focus on building the user interface rather than dealing with complex setup processes.

For more detailed information about the homepage, please refer to the official repository available at Swarm-in-Blocks Repository.

*C. First Person View (FPV)*

The development of the Swarm First Person View (sFPV) system allows drone pilots to control flights in real time through a camera installed on the drone. This feature has been restructured to run entirely on the web, integrating seamlessly with the Swarm Station. The sFPV system enhances drone operation by providing real-time visuals and data, such as battery levels, telemetry, CPU status, and flight state, thereby improving flight safety and control.

*1) Code Structure:* The `sFPV` package is organized into two main parts: the backend and the frontend. Each plays a critical role in ensuring the system's functionality:

*a) `msg/`:* This directory contains the messages used to handle user requests and manage communication within the system. These messages are essential for relaying commands and responses between the frontend interface and the backend services.

*b) `src/`:* The `src/` directory houses the core script files needed to run the FPV application. These scripts are responsible for capturing the drone's camera feed, processing it, and relaying it to the user interface for real-time display.

*c) `launch/`:* The `launch/` directory contains the necessary launch files to execute the FPV system. These files allow users to initiate the application from the terminal, setting up the environment and connecting the necessary ROS nodes for the operation of sFPV.

*2) Frontend:* The frontend of the FPV application is built using `roslib.js` to facilitate communication with the ROS backend. The web interface, developed with HTML and CSS, allows users to control the drone through their browser. This interface displays the live video feed from the drone's camera, along with essential telemetry data, enabling precise and safe drone operation. The user can interact with the drone using computer keyboard inputs, adding another layer of control directly from the web interface.

Fig. 8. First Person View Code.

For more detailed information about the sFPV, please refer to the official repository available at Swarm-in-Blocks Repository.

*D. Swarm Station*

The `swarm_station` package is responsible for running the backend of the Swarm Station, which provides a fully integrated control center for managing drone swarms. This package utilizes the `tf2_web_republisher` package alongside the `roslibjs` and `ros3djs` libraries to serve a 3D web visualizer. The system runs almost entirely on the web, allowing users to access the Swarm Station from any device connected to the same network as the host.

*1) Code Structure:* The `swarm_station` package is organized into several key directories, each playing a crucial role in the overall functionality of the Swarm Station:

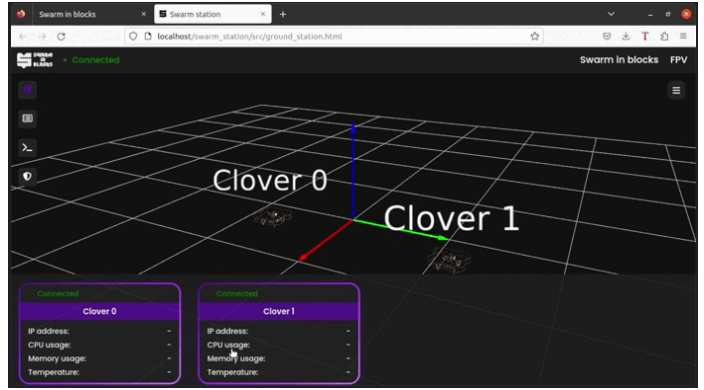

Fig. 9. Swarm in Blocks Swarm Station.

*a) `launch/`:* The `launch/` folder defines and initializes nodes, services, and parameters necessary to start the application. It also includes specific configurations required for the package.

*b) `meshes/`:* This directory contains definitions of the physical and dynamic characteristics of the Clover's 3D model used for simulation purposes.

*c) `src/`:* The `src/` directory is the core of the package, where source code is developed and compiled. It includes the definitions responsible for creating nodes, services, and other functionalities related to the frontend of the Swarm Station.

*2) Main Features:* The Swarm Station includes several integrated features designed to streamline swarm management:

*a) Information Center:* A central hub for displaying essential data on each drone, including an emergency "land all" button for quick response.

*b) Drone's Process:* Displays hardware information from each Clover's Raspberry Pi, making it easier to identify and compare potential issues.

*c) Topic List:* A feature that lists currently active topics, allowing users to monitor and analyze ongoing processes without needing an additional terminal.

*d) Web Terminal:* A web-based terminal that allows users to send commands directly from the web interface, avoiding the need for multiple screens. Multiple terminals can be opened simultaneously for handling various processes.

*e) Safe Area:* A feature that lets users define a safe operating area for the drones. If a drone exits this area, it is automatically landed, enhancing safety.

For more detailed information about the Swarm Station, please refer to the official repository available at Swarm-in-Blocks Repository.

## VIII. COMMUNICATION GRAFICS

The Swarm-in-Blocks platform effectively integrates MAVROS, ROS, and its web-based interface to manage and control multiple drones. MAVROS acts as a bridge between the drone's flight controllers and the ROS ecosystem, translating drone-specific commands and telemetry into a ROS-compatible format.

ROS serves as the central hub, coordinating the flow of information between the drones and the user interface. The Swarm-in-Blocks interface, built using `roslibjs`, communicates with ROS nodes in real-time, allowing users to control drones through an intuitive block-based programming environment.

Commands created in the interface are converted into ROS commands and sent to MAVROS, which executes them across the connected drones. The drones' telemetry and status updates are fed back through this communication chain, ensuring that users can monitor and adjust operations as needed.

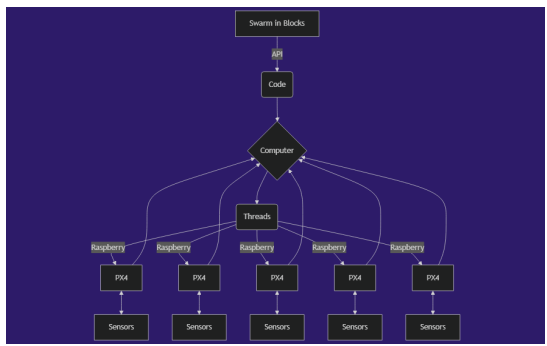

Fig. 10. Schematic of communication flow between MAVROS, ROS, and Swarm-in-Blocks.

This streamlined communication framework ensures cohesive and synchronized control of the entire drone swarm.

## IX. SIMULATION ANALYSIS

Simulation analysis is essential for validating and optimizing robotic systems before real-world deployment. By using advanced tools like Gazebo, we can recreate complex scenarios, evaluate system performance, and identify potential issues in a virtual environment. This section focuses on integrating two systems within Gazebo, covering the setup process, system interaction, and outcome analysis to refine design and functionality.

This method provides a thorough evaluation of system behavior under various conditions, allowing developers to enhance reliability and efficiency. Whether testing a single robot or a complex multi-agent scenario, the insights from these simulations are crucial for guiding development and ensuring successful deployment.

### A. The Systems

The first system features an Nvidia RTX 3060 mobile GPU, an Intel i7-11800H processor, and 16GB of DDR4 RAM, offering a strong blend of graphical and computational power for smooth, high-resolution simulations in Gazebo. The second system, with an AMD Ryzen 5 3600 CPU, a Radeon RX 5700 XT OC GPU, and 16GB of DDR4 RAM, provides robust processing and excellent graphical rendering, making it well-suited for intensive simulation tasks requiring both CPU and GPU performance.

### B. Real Time Factor

The Real-Time Factor (RTF) in Gazebo serves as an indicator of how efficiently the simulation runs compared to real-time. Ideally, an RTF of 1.0 signifies that the simulation is running exactly in real-time, while an RTF below 1.0 indicates that the simulation is running slower than real-time, typically due to increasing computational demands.

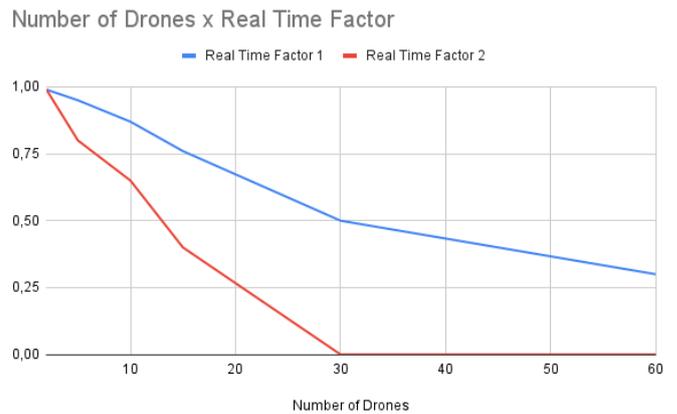

Fig. 11. Comparisson of RTF between 2 systems.

In the graph provided, the blue line represents the performance of the AMD-based system (AMD Ryzen 5 3600 with Radeon RX 5700 XT OC), while the red line represents the Nvidia-based system (Intel i7-11800H with Nvidia RTX 3060 mobile). As the number of drones increases, the RTF decreases for both systems, reflecting the higher computational load.

However, a critical observation is that when the number of drones exceeds 10, the RTF for the Nvidia-based system (red line) drops precipitously and eventually crashes the system, as indicated by the RTF reaching near zero. This crash highlights the Nvidia-based system's inability to handle the simulation workload as effectively as the AMD-based system, which, although also experiencing a decline in RTF, manages to sustain the simulation for a larger number of drones. This suggests that the AMD system provides better stability and performance under high-load conditions in Gazebo, whereas the Nvidia system struggles and ultimately fails when the number of drones increases beyond 10.

### C. Analysis

The scalability of the system is directly influenced by the performance of its central node. The AMD-based system (Ryzen 5 3600 with Radeon RX 5700 XT OC) demonstrates better scalability, as it can manage an increasing number of drones without crashing, making it suitable for expanding the simulation's complexity. In contrast, the Nvidia-based system (i7-11800H with RTX 3060 mobile) shows limited scalability, struggling with more than 10 drones, which leads to system crashes and hinders the ability to scale the simulation effectively.

To achieve greater scalability, upgrading to more powerful and modern components would enhance the system's capacity to handle larger simulations and real life number of drones. This would ensure that the central node maintains performance and stability as the number of drones and the complexity of the simulation increase, allowing for a more scalable and robust solution in demanding environments.

## X. RESULTS AND DISCUSSION

The Swarm-in-Blocks framework has shown significant improvements in drone swarm coordination and ease of use. Its block-based interface allows for quick prototyping and iterative testing, enabling users with minimal programming experience to manage drone swarm operations effectively [3].

### A. Scalability and Flexibility

Swarm-in-Blocks is highly scalable, supporting large numbers of drones without added complexity. The modular blocks offer flexibility in defining and modifying swarm behaviors, essential for large-scale applications like environmental monitoring and disaster response [4].

### B. Safety and Reliability

Decentralized control mechanisms enhance the system's safety and reliability. Autonomous operation and redundancy ensure the swarm continues functioning even with drone malfunctions [5].

### C. Performance Evaluation

Swarm-in-Blocks handles real-time operations with minimal latency. The block-based interface simplifies programming, allowing users to focus on high-level strategies. Benchmark tests confirm efficient performance across various environments [3].

### D. User Feedback

Users have praised the intuitive design and ease of use. Educational institutions find it particularly valuable for teaching swarm robotics, bridging the gap between theory and practice [4].

## XI. FUTURE WORK

Swarm-in-Blocks is continuously evolving, with future work focusing on:

### A. Integration with Other Platforms

Integrating with other robotic platforms for seamless interoperability, enabling comprehensive solutions for complex tasks [3].

### B. Enhanced User Interface

Improving the user interface with more customization, real-time feedback, and monitoring tools to simplify large swarm management [4].

### C. Field Testing and Validation

Conducting extensive field testing in diverse environments to ensure robustness and identify areas for improvement [5].

## XII. CONCLUSION

The Swarm-in-Blocks project simplifies the programming of drone swarms using a block-based language, making it accessible to users with limited technical knowledge. The Swarm-in-Blocks framework provides a user-friendly and scalable platform for communication and coordination, with potential applications in areas such as environmental monitoring, disaster response, industrial automation, agriculture, and entertainment. Future work will focus on further optimizing the framework and exploring additional applications in various domains.

ACKNOWLEDGMENTS TO THE ATENA TEAM

We would like to express our deep gratitude to the Atena team, part of the SEMEAR Group (Solutions in Mechatronic Engineering and Robotics Application) at the University of São Paulo (USP), São Carlos campus.

Comprising over 100 members, including undergraduate and graduate students, and supported by professors, the Atena team participated in competitions such as CopterHack. Team members include Agnes Bressan de Almeida, João Aires Correa Fernandes Marsicano, Felipe Andrade Garcia Tommaselli, Gabriel Ribeiro Rodrigues Dessotti, José Carlos Andrade do Nascimento, Lucas Sales Duarte, Matheus Della Rocca Martins, and Nathan Fernandes Vilas Boas. Their dedication and collaboration were essential for the project's success.

For more information, visit the SEMEAR website.